# Generating Privacy-Preserving Process Data with Deep Generative Models

Keyi Li[1], Sen Yang[2], Travis M. Sullivan[3], Randall S. Burd[3], Ivan Marsic[1]
[1]Electrical and Computer Engineering Department, Rutgers University, [2]LinkedIn, [3] Children's National Hospital
{kl734, sy358, marsic}@rutgers.edu, {tsullivan, rburd}@childrensnational.org

## ABSTRACT

Process data with confidential information cannot be shared directly in public, which hinders the research in process data mining and analytics. Data encryption methods have been studied to protect the data, but they still may be decrypted, which leads to individual identification. We experimented with different models of representation learning and used the learned model to generate synthetic process data. We introduced an adversarial generative network for process data generation (ProcessGAN) with two Transformer networks for the generator and the discriminator. We evaluated ProcessGAN and traditional models on six real-world datasets, of which two are public and four are collected in medical domains. We used statistical metrics and supervised learning scores to evaluate the synthetic data. We also used process mining to discover workflows for the authentic and synthetic datasets and had medical experts evaluate the clinical applicability of the synthetic workflows. We found that ProcessGAN outperformed traditional sequential models when trained on small authentic datasets of complex processes. ProcessGAN better represented the long-range dependencies between the activities, which is important for complicated processes such as the medical processes. Traditional sequential models performed better when trained on large data of simple processes. We conclude that ProcessGAN can generate a large amount of sharable synthetic process data indistinguishable from authentic data.

## 1 Introduction

Process mining (PM) techniques have been applied to discover knowledge about processes in different disciplines, including healthcare [20], business [25], and education [3]. Process data in these fields usually contains confidential information, adding a requirement for addressing privacy concerns and obstacles related to data sharing [23]. Public process datasets are limited compared to other data types such as images, text, and sensor. Data sharing has led to advancement in many research fields (e.g., ImageNet in computer vision and GLUE in natural language processing) by standardizing the research practice and enabling a focus on model exploration and benchmark comparisons. The lack of data is a barrier of process mining studies in many domains. This work addresses the challenge of sharing process data with a privacy-preserving requirement. The principle is to first learn the underlying data distribution by learning representative deep sequential models (e.g., gated recurrent unit (GRU) [5], long short-term memory networks (LSTM) [10], Transformer [26]), and to use these models to generate high quality synthetic process data (Figure 1). The synthetic data generated will have a similar distribution as actual data and can be shared without compromising confidentiality. In addition, the size of the process data can be increased to include novel process sequences.

Process data is sequential data constructed from a series of activities. Process data varies according to use cases. Simple processes can be almost linear (i.e., single path) with only a few activities, while complex processes can have parallel paths, loops, intermediate dependencies, and hundreds of different activities.

The recurrent neural networks (RNN), such as GRU and LSTM, have been successfully applied to learning sequential data [21]. For long and complex sequences, the RNNs may have drawbacks. First, the inner dependencies of the tokens in these sequences are difficult to capture by the RNNs. Second, when the training data size is small, convergence of the RNNs can be difficult. These factors may impact the reliability of the synthetic process data generated by RNNs.

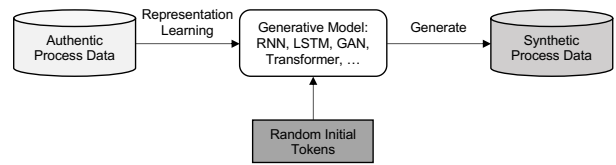

**Figure 1: Our framework of generating synthetic process data by representation learning from authentic process data.**

We aim to explore different methods for synthetic sequence generation that address different degrees of complexity. We focused on (1) generating plausible synthetic process data with few process errors, (2) learning the authentic data distribution when data size is small, (3) generating synthetic process data with only random initial tokens as inputs to preserve the privacy of authentic data, and (4) generating new process sequences that are plausible based on the actual process. Inspired by the image generation with random noises, we applied the Generative Adversarial Networks (GANs) for process data generation [6]. We propose ProcessGAN, built based on the encoder part of Transform-er and a GAN structure, to generate process data[1]. The Transformer model uses the *multi-head self-attention* mechanism to address long-term dependencies, while the GAN structure increases the precision of the synthetic sequences when data size is insufficient. Unlike the RNNs that take one initial token as input to generate a sequence step-by-step, the Transformer model can take a full random sequence as input. The ProcessGAN model can then expand the exploration space and generate more variable process sequences.

We evaluated the synthetic data quantitatively by measuring the underlying distributions and qualitatively by assessing the semantics. We calculated basic statistical measures between the synthetic data and the authentic data to compare the performances of different models. We used both a data-driven method and a knowledge-based assessment to evaluate the semantics. We did a case study by inviting domain experts to check the process errors occurred in the synthetic process dataset.

The main contributions of this paper are:
- A framework for generating privacy-preserving process data. We compared different models and analyzed their merits.
- A novel GAN-based Transformer network to generate realistic process data. In this model, we leveraged Gumbel-Softmax to handle the discrete backpropagation from the discriminator and added activity distribution into the loss function to achieve better generation performance.
- A solution for evaluating the synthetic process data. The solution includes three different dimensions: (1) statistical measures, (2) supervised learning score, and (3) process mining.

## 2 Related Work

### 2.1 Process Data Generation

Process data can be difficult to collect and share because of complexity and inclusion of private information [27]. The hidden Markov model has been used to augment process datasets to capture the sequential dependencies of activities and generate synthetic process data [29]. In recent years, some deep generative models, such as the RNN-related models, Transformer model, and GANs, were applied to the process mining field to do the process events prediction task [2, 16, 24]. These models can also be used to generate process data. The difference is that the prediction task usually applies the teacher forcing mechanism to do the training and prediction (i.e., using the activities in the ground truth sequences to predict the next activity). However, in the data generation task, we do not have access to the ground truth sequences, and the next token is generated based on the previous generated tokens. In the case of some long complex processes, RNNs may suffer exposure bias [1, 19], i.e., errors occurring in the generated tokens propagate during data generation. These errors may reduce the authenticity of synthetic process data. The Transformer network can manage long-term dependencies using the self-attention mechanism, but the autoregressive training (i.e., generating the next step based on the previous generated steps) of the Transformer is not computational efficient. We sought to address these challenges by developing a more efficient non-autoregressive training (i.e., generating all tokens in a sequence in parallel) of the Transformer and adding the GAN structure for better generation performance.

### 2.2 Generative Adversarial Networks

Our idea of generating sharable synthetic data was inspired by the concept of image data augmentation in machine learning tasks [22]. GAN-based neural networks have been trained on limited authentic data to enhance data size while capturing the features and generating plausible images using only random noises [6]. When using GAN, a discriminative network $D$ was trained to distinguish between authentic and generated synthetic data, while a generative network $G$ was trained to generate synthetic data from random input and tricks $D$. $G$ gradually learned to mimic the features of the authentic data and learned to generate synthetic data that is similar to actual data. This high-quality synthetic data can be used to augment training dataset for machine learning and can be shared preserving privacy.

GAN-structured generative networks have also been applied for text data generation [9, 17, 30-32]. In these works, generators are usually RNN-based models, and discriminators are usually Convolutional Neural Network (CNN)-based or RNN-based models. Because the gradients from the discriminator cannot back-propagate through discrete variables, SeqGAN was introduced that use reinforcement learning (RL): each time a new token is generated, $G$ simulates the whole sequence and lets $D$ score the sequence, then $D$ gives an RL reward to that token [11, 30]. With the RL reward given by $D$, some error tokens will have lower scores and will be less likely to be generated. $G$ then learns the intermediate dependencies of the tokens and generates high-quality sequences.

Although the RL rewards given by the GAN discriminator can reduce some errors, the training of these GANs is difficult. The discriminator's high variance of RL rewards may confuse the generator. The training process may experience mode collapse because the generator cannot learn a stable representation from the discriminator [14]. These problems are obstacles for sequential data generation, especially for complex process data.

## 3 Method

In this section, we introduce the ProcessGAN model and methods for evaluating the synthetic process traces ("sequences" and "traces" are used alternatively in the following sections). In ProcessGAN (Figure 2), we use the Transformers (encoder only) as both the generator $G$ and discriminator $D$ of the GAN architecture. The input to the generator is a random sequence, and the output from the generator is a sequence of vectors. Each dimension of a vector has each activity's probability. We then extracted the most likely activities from the vectors and built a synthetic activity sequence. Considering that this extraction is non-differentiable, we applied straight-through Gumbel-Softmax to obtain the gradients to update the parameters of the generator. The synthetic and authentic process sequences are then fed into the discriminator for adversarial learning. We included an auxiliary loss that considers the activity distribution divergence to guide adversarial learning. We propose several evaluation approaches of different perspectives, like data statistics, supervised learning, and process mining.

---
[1] The source code is available at https://github.com/raaachli/ProcessGAN

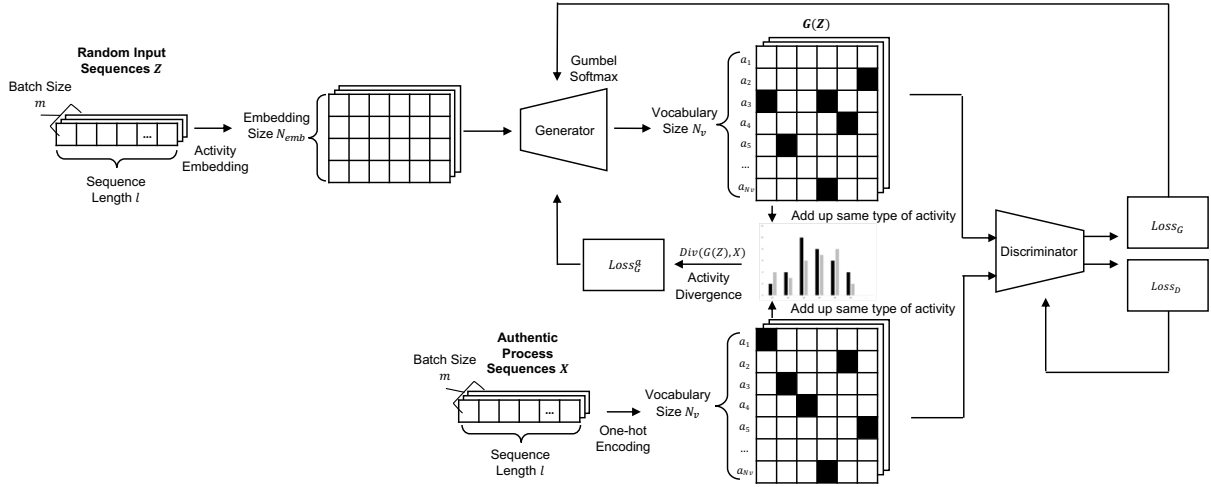

Figure 2: The architecture of the ProcessGAN network. The random input sequences and authentic process data (authentic activity sequences) are preprocessed to the same length. Discriminator is trained to distinguish the authentic and synthetic sequences. Generator is optimized by the losses given by the discriminator and the activity divergence.

### 3.1 ProcessGAN Model

**Transformer and GAN.** Traditional Maximal Likelihood Estimation (MLE) training of the RNN-related networks suffers from exposure bias, i.e., RNNs tend to overfit the exposed data and would not explore the vast output space beyond the exposed data distribution [19]. If a sequence generated from an unobserved starting token, divergence may occur and become enlarged along the autoregressive generation process because each token is predicted based on previously predicted tokens.

Unlike the RNN-related models where the current token are highly depends on previous tokens, the Transformer network uses the positional encoding technique to inform the relative positions of the tokens [26]. For this reason, the generation is more robust to the initial inputs and can generate all tokens of a sequence in parallel. This non-autoregressive step alleviates the accumulated discrepancies when tokens are generated one by one and can handle longer, more complex sequences. In addition, with adversarial training, feeding Transformer with random sequences as input can regularize the entire model and reduce exposure bias.

**Data Preparation.** For activities $\{a_i\}$ in the process dataset, we build an activity vocabulary $V$ (vocabulary size as $N_v$) using the index-based encoding. For each authentic process sequence $x = [a_1, a_2, \ldots a_n]$, we pad a predefined end token $[end]$ to the end of the sequence when the length is shorter than the maximal sequence length $l$. Random input sequence $z = [a_1, a_2, \ldots a_l]$ is generated by random sampling of the activity vocabulary (including the end token) until the max sequence length is reached.

Our experiment shows the end token in the random sequence will not affect the result.

**Generator.** The generator $G$ takes random sequences $Z$ as input. We first embed the activities in $Z$ (like word-embedding in NLP). We set the embedding size ($N_{emb}$) according to a rule of thumb: $N_{emb} = \sqrt[2]{N_v}$ (in NLP the rule of thumb is $\sqrt[4]{N_v}$. For process data, the activity vocabulary size is usually smaller, so a good start point is $\sqrt[2]{N_v}$).

The Transformer network uses the *self-attention* mechanism to learn each token's position and correlations with other tokens. The rich dependencies between the activities in a process (i.e., one activity could have multiple relations with other activities) were captured based on the multi-head self-attention mechanism. After $M$ self-attention blocks and a linear layer, the output sequences are represented by categorical vectors (denoted as $\boldsymbol{h}$) with $N_v$ dimensions, where each vector represents a probability distribution over the activities in the vocabulary list.

We applied argmax on $\boldsymbol{h}$ to obtain one-hot representation of output sequences, where: $\boldsymbol{s}_i = [\boldsymbol{o}_1, \boldsymbol{o}_2, \boldsymbol{o}_3, \ldots, \boldsymbol{o}_l]$, $\boldsymbol{o}_i = onehot(\arg\max \boldsymbol{h}_i))$.

**Discriminator.** For discriminator $D$, we also use the Transformer encoder to avoid mode collapse that may cause by the learning rate mismatch between generator and discriminator during adversarial learning. We find the first end token for the generated sequence and pad all the tokens after it. $D$ takes sequences (generated and authentic) of one-hot vectors as input and outputs a binary classification result.

**Learning Objectives.** The generator and discriminator were optimized alternatively to achieve an equilibrium. The discriminator was used to score the generated sequences to train the generator. The generator's objective is to maximize this score.

The output of the last layer of the generator is the sequences of vectors $\boldsymbol{h}$, with each dimension of $\boldsymbol{h}$ representing the probability of that token to be generated in the position. In the for-ward propagation, we applied argmax on $h$ to generate a sequence in one-hot encoding representation, which is then fed into the discriminator. The argmax operation is not differentiable in the back propagation, and the gradient cannot be passed back. We adopted the straight-through Gumbel-Softmax mechanism with a differentiable sampling operation [11]. The discriminator scores the generated sequences and passes the gradient back to the generator. The loss function of generator is:

$$\mathcal{L}_G = -\mathbb{E}_{Z \sim P(Z)} \log D(G(Z)) \quad (1)$$

where: $G(Z) = S = [s_1, s_2, s_3, ..., s_m]$.

We encoded the authentic sequences $X$ into one-hot representation to make them the same expression as the generated sequences. The discriminator distinguishes between the generated and the authentic sequences. The loss function of discriminator is:

$$\mathcal{L}_D = -\mathbb{E}_{X \sim P_{data}} \log D(X)$$
$$- \mathbb{E}_{Z \sim P(Z)} \log(1 - D(G(Z))) \quad (2)$$

Compared with RNNs, the random inputs and adversarial training in ProcessGAN help reduce exposure bias. Due to the larger search space, it is harder for the ProcessGAN to converge [8]. To help reduce the search space, we added activity distribution divergence as an auxiliary loss to the generator. The activity distribution divergence was defined in two different ways, Kullback-Leibler (KL) divergence and mean squared error (MSE):

$$\mathcal{L}_G^a = KL_{div}(X'_a || S'_a) = \frac{1}{m} \sum_{i=1}^{N_v} S'_a(a_i) \log\left(\frac{X'_a(a_i)}{S'_a(a_i)}\right) \quad (3)$$

$$\mathcal{L}_G^a = MSE(X'_a, S'_a) = \frac{1}{m \times N_v} \sum_{i=1}^{N_v} [X'_a(a_i) - S'_a(a_i)]^2 \quad (4)$$

where $m$ is the batch size, $X'_a$ and $S'_a$ are the activity probability distributions of one batch of authentic sequences and generated sequences. We calculated the divergence in each training batch and feed it to generator (Figure 2). With divergence loss ($\mathcal{L}_G^a$), the generator loss becomes:

$$\mathcal{L}_G' = \mathcal{L}_G + w_a \mathcal{L}_G^a \quad (5)$$

where $w_a$ is the weight of the auxiliary losses.

$w_a$ is a hyperparameter that balances adversarial loss and divergence loss. A heuristic choice of $w_a$ is a value that enforces the two losses of similar scale. We first run the generator several times without optimization and obtain the expectations of adversarial loss $\mathbb{E}(\mathcal{L}_G)$ and divergence loss $\mathbb{E}(\mathcal{L}_G^a)$. $w_a$ can be approximated to $\frac{\mathbb{E}(\mathcal{L}_G)}{\mathbb{E}(\mathcal{L}_G^a)}$.

**Adversarial Training.** At the early adversarial training stage, the $D$ may converge quickly because $G$ is generating implausible sequences. $D$ may then reject all the generated sequences and lead to poor optimization of $G$. To control the optimization speed and slow down $D$'s convergence, we adopted the training scheme of optimizing $k$ epochs ($k$ set to 2 in our experiments) of $G$ and one epoch of $D$ [6].

### 3.2 Evaluation Methods

Evaluation of sequential data is not as intuitive as image data. Synthetic images are usually evaluated based on the authenticity of characters or the resolution, which is easy to observe [6, 12]. We aim to generate synthetic process data that follows the underlying distribution of the real-world process where authentic data is observed. Our evaluation takes a sample of synthetic sequences and evaluates using three measures, i.e., statistical measures, supervised learning score, and workflow discovery.

**Statistical Measures.** We use three measurements to measure the distance between synthetic and the authentic sequences.

(a) Sequence length: the number of activities in a sequence. We compared the mean and standard deviation of sampled synthetic and authentic sequences.

(b) Activity type occurrence: the count of occurrence of each activity type. Because each process data has different number of activity types, we reported an aggregated metric to measure the difference of activity occurrence:

$$Distance(X_a, S_a) = \sum_{i=1}^{N_v} |X_a(a_i) - S_a(a_i)|$$

where $X_a, S_a$ represent the activity frequency distribution of the datasets. The metric measures the L1 distance between the percentages of each activity type. We used L1 distance here for better interpretability instead of KL divergence.

(c) Sequence variance: the variance of sequences measures the spread or distance between sequences in a dataset. We used the *sum of pairwise normalized edit distance* (SPE) to measure variance:

$$SPE = \frac{1}{N^2} \sum_{i=1}^{N} \sum_{j=i+1}^{N} ED(s_i, s_j) / (\text{length}(s_i) + \text{length}(s_j))$$

where $N$ is the number of sequences in the generated samples, $s$ are the sequences and $ED$ is the edit distance (Levenshtein distance).

These three measurements describe the basic dimensions of a process sequence. The activity type occurrence is a cross measurement, i.e., computed across synthetic and authentic process sequences. The smaller distance indicates higher affinity between synthetic data and authentic data.

**Supervised Learning Score.** We can invite domain experts to manually evaluate the generated sequences. This manual assessment is labor-intensive and suffers from human bias. For this reason, we trained an independent binary classifier and used it as an "off-the-shelf" supervised scorer to assess the realism of the synthetic data. The classifier can be an evaluator with comprehensive domain knowledge that can distinguish very similar positive and negative sequences. We adopted the Transformer model as the binary classifier. Traditional sequential classification models like RNN, CNN, or tree models with crafted features as input, can also be used as the classifier.

Similar to the negative sampling method in the NLP field, we created high-quality activity sequences on the negative samples by manually infusing noise (randomly adding, deleting, and switching a predefined ratio of tokens) to the authentic sequences [15]. To distinguish more features of the data, after the Trans-former layer we concatenated each sequence's output with its activity frequency distribution and sequence length and sent them to two additional dense layers (Figure 3). Let $\theta$ be the parameters of the model, the objective is to maximize the log-likelihood w.r.t $\theta$:

$$\theta \to \sum_{t \in \mathbb{T}} \log p(y|t, \theta)$$

where $\mathbb{T}$ is the training set that contains the positive and negative samples, and $y$ is the correct class of $t$.

We generated negative samples five times more than the positive samples to augment the training data. After the Transformer achieved a good F1 (F1>0.8), we used it to score the synthetic sequences.

**Process Mining.** We applied the workflow discovery method from the process mining field to construct a workflow diagram from the generated process traces [28]. We aimed to visualize the process traces and evaluate whether they represented the underlying distribution of authentic data.

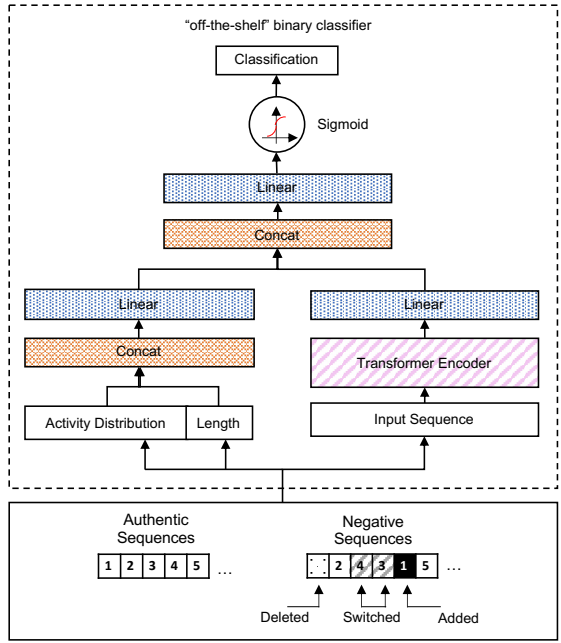

**Figure 3: To generate negative sequences, we randomly add noises to the authentic sequence. The input of the classifier is trained based on the sequences, their activity distribution and length.**

To construct the diagram, we used the trace alignment method to generate consensus sequence from the alignment result [4]. A consensus sequence captures the major activities in the process traces and can be considered the workflow diagram's backbone. Some activities that are not in the consensus sequence but are required are added as the side branches in addition to the backbone to represent the parallel activities, and some infrequent activities are filtered out [13]. We asked domain experts to compare the workflow diagrams and check for process errors.

## 4 Experiment and Results

### 4.1 Dataset Introduction

We performed our experiment and analysis on four medical process datasets of pediatric trauma resuscitation (the secondary survey process, intubation process, emergency department procedures, and maintaining airway process). The process traces were manually coded from videos. The use of these datasets was approved by the Institutional Review Board of Children's National Hospital in Washington, DC. In addition, we used two public process datasets[2] to evaluate our method on different types of processes: a real-world event log which contains the process of sepsis cases from a hospital, and an event log of a loan application process (Table 1). The loan application process has a large length variance. To test our model for generating process with different degrees of complexity, we split this dataset into two parts: one with sequence lengths $\leq$ 50 and the other > 50.

### 4.2 Experimental Design

**Baseline Methods.** We used three traditional generative networks for creating sequential data as baselines for evaluating of our ProcessGAN network: GRU, LSTM, and autoregressive Transformer (Trans-AR). These networks sequentially generated synthetic process traces.

**Table 1: Statistics of four medical datasets and two public datasets, including the case number, mean value and standard deviation of the sequence length, and the number of activity types. (D1: Secondary survey process of trauma resuscitation; D2: Intubation process of trauma resuscitation; D3: Emergency department process; D4: Maintain airway process of trauma resuscitation; D5: Sepsis process; D6: Loan application process dataset, where the first part contains the sequences with length $\leq$ 50 and the second part contains the sequences with length > 50.)**

|  | D1 | D2 | D3 | D4 | D5 | D6_1 | D6_2 |
|---|---|---|---|---|---|---|---|
| # Cases | 271 | 101 | 381 | 53 | 846 | 3271 | 1013 |
| Length (Mean) | 46.03 | 12.35 | 5.53 | 31.72 | 17.28 | 34.50 | 68.89 |
| Length (Std.) | 15.20 | 3.18 | 2.13 | 18.08 | 12.01 | 8.58 | 18.03 |
| # Act. Types | 47 | 15 | 55 | 38 | 17 | 37 | 37 |

**Our Methods.** We first tried a non-autoregressive Transformer to generate sequences using random sequences as input without GAN structure. We then used the proposed ProcessGAN model and two of its variants to generate synthetic process traces.

- Trans-NAR: A non-autoregressive Transformer network that uses random sequences with fixed length as input and generates synthetic process traces in parallel.
- P-GAN: The vanilla ProcessGAN model with adversarial training only. It is a GAN-based Transformer network that uses random sequences with fixed length as input and generates synthetic process traces in parallel.
- P-GAN-M: P-GAN variant with the mean squared error loss as activity distribution loss.

---
[2] http://www.processmining.org/event-data.html

Table 2: The sequence lengths of the process data generated by different models.

|  | D1 | D2 | D3 | D4 | D5 | D6_1 | D6_2 |
|---|---|---|---|---|---|---|---|
| Authentic | 46.89±12.46 | 13.3±2.69 | 5.68±2.07 | 31.0±23.49 | 17.53±9.50 | 34.85±8.71 | 67.87±17.45 |
| GRU | 42.33±22.15 | 11.37±3.46 | 5.20±2.14 | 28.27±20.43 | 15.78±7.42 | 32.88±9.60 | 70.87±34.05 |
| LSTM | 44.68±21.02 | 11.41±2.87 | 5.21±2.11 | 20.18±16.47 | 15.82±8.36 | 32.18±9.10 | 68.17±33.76 |
| Trans-AR | 39.38±18.27 | 11.36±2.86 | 5.08±1.92 | 25.89±14.06 | 16.71±7.59 | 33.25±8.81 | 56.56±26.80 |
| Trans-NAR | 31.53±5.38 | 11.27±1.64 | 4.56±1.41 | 12.90±4.14 | 11.97±2.68 | 24.86±4.86 | 56.68±6.05 |
| P-GAN | 43.79±13.32 | 12.40±1.76 | 5.31±1.92 | 33.09±15.91 | 17.19±10.37 | 33.78±7.98 | 63.29±15.57 |
| P-GAN-M | 43.49±11.79 | **12.70±1.69** | **5.74±1.76** | 35.82±9.54 | **17.25±8.47** | **34.84±7.82** | **68.62±19.87** |
| P-GAN-K | **46.24±18.13** | 12.21±2.02 | 5.58±1.64 | 29.71±22.57 | 17.10±5.18 | 33.59±9.48 | 64.59±6.61 |

- P-GAN-K: P-GAN variant with the KL divergence loss as activity distribution loss.

The baseline models are all autoregressive models. These models were trained through the authentic sequences by using the ground truth tokens to predict the next token in a sequence, i.e., $\prod_{i=1}^{T} P(a_{i+1}|a_i, a_{i-1}, ..., a_1, a_0)$. For the Trans-NAR model, we stopped training when the loss converged. It can be difficult for GAN models to tell the stop point, especially for sequential data that are not as intuitive as images. The loss value of both $G$ and $D$ were oscillating because they were playing a minimax game. Ideally, the accuracy of $D$ will first rise because $G$ is poor. As the performance of $G$ gets better, $D$'s accuracy will drop. If the accuracy is oscillating around 0.5, $G$ is generating plausible sequences to tricks $D$. For this reason, we picked the result when $D$ and $G$ reach this equilibrium [6].

We divided each dataset into training, validation, and testing sets in a 0.8:0.1:0.1 ratio. For each model, we generated 500 synthetic sequences and compared them with the hold-out test set.

Table 3: The overall activity type occurrence differences of the sequences generated by different models.

|  | D1 | D2 | D3 | D4 | D5 | D6_1 | D6_2 |
|---|---|---|---|---|---|---|---|
| GRU | **0.11** | **0.16** | 0.23 | 0.64 | 0.09 | **0.10** | 0.08 |
| LSTM | **0.11** | 0.17 | 0.24 | 0.68 | 0.08 | 0.14 | **0.06** |
| Trans-AR | 0.19 | 0.17 | **0.20** | 0.58 | 0.11 | 0.13 | 0.10 |
| Trans-NAR | 0.24 | 0.18 | 0.31 | **0.55** | 0.32 | 0.37 | 0.17 |
| P-GAN | 0.31 | 0.23 | 0.45 | 0.77 | 0.15 | 0.34 | 0.32 |
| P-GAN-M | 0.17 | 0.21 | 0.29 | **0.55** | **0.07** | 0.22 | 0.17 |
| P-GAN-K | **0.11** | **0.16** | 0.23 | 0.60 | 0.11 | 0.11 | 0.28 |

### 4.3 Results

#### 4.3.1 Quantitative results

**Statistics.** We repeated each experiment several times and found the results are stable. For the sequence lengths (Table 2), we found the ProcessGAN-based models outperformed other models without a GAN structure, especially for the datasets with longer sequences. The discriminator of the GAN networks can help the generator avoid generating biased sequences that are either too long or too short. For longer sequences with smaller data sizes (D1, D4, D6_2), we observed that the RNN-based models and the autoregressive Transformer model tend to generate sequences with shorter lengths or a larger standard deviation (Figure 4). For shorter sequences (D2, D3) or sequences with relatively sufficient data size (D5, D6_1), the autoregressive models can generate comparable sequence lengths. We also observed the non-autoregressive Transformer model generated shorter sequences with smaller standard deviations on all the datasets. Non-autoregressive Transformer model was trained by minimizing the cross-entropy loss between the full generated sequences and the authentic sequences directly. It generated tokens with the assumption of independence, which is not reasonable for sequential data generation [7]. The model usually converges faster than other models but fails to generate precise sequences. The GAN-based model that incorporates an additional discriminator can help the generator find sequences with more precise activity orders, which addresses the independence assumption problem.

Table 4: The sequence variance (SPE) of authentic data and synthetic data generated by different models. Synthetic sequence variance that is closer to authentic data is considered to be better.

|  | D1 | D2 | D3 | D4 | D5 | D6_1 | D6_2 |
|---|---|---|---|---|---|---|---|
| Authentic | 0.21 | 0.14 | 0.21 | 0.23 | 0.17 | 0.16 | 0.17 |
| GRU | 0.28 | 0.19 | 0.24 | 0.31 | **0.17** | **0.17** | 0.21 |
| LSTM | 0.27 | **0.17** | 0.24 | 0.32 | 0.18 | **0.17** | 0.21 |
| Trans-AR | 0.29 | 0.20 | 0.25 | 0.29 | **0.17** | 0.18 | 0.23 |
| Trans-NAR | **0.24** | 0.19 | 0.24 | 0.25 | 0.16 | 0.19 | 0.20 |
| P-GAN | **0.24** | 0.18 | 0.23 | 0.26 | **0.17** | 0.19 | **0.18** |
| P-GAN-M | 0.25 | 0.18 | **0.22** | **0.23** | **0.17** | **0.15** | 0.19 |
| P-GAN-K | 0.26 | 0.19 | **0.22** | 0.26 | 0.15 | 0.20 | **0.18** |

We found that the autoregressive models perform better for activity type occurrence (Table 3). This observation is expected because the models are trained with the MLE objective through teacher forcing, in which case the activity distribution of the training data can be easily learned. Vanilla ProcessGAN shows a large difference in activity type occurrence with authentic data. With activity distribution divergence as an auxiliary loss, P-GAN-M and P-GAN-K also generated comparable activity distri-bution to the autoregressive models (Table 3, Figure 5). For complex processes (i.e., longer sequence and large activity vocabulary) with very limited data size (D4), the autoregressive models did not have an advantage in generating a good activity distribution. The performance of these models relied on sufficient data.

Table 5: The fraction of the synthetic sequences that the "off-the-shelf" binary classifier classified as authentic. This fraction measures how frequently the synthetic data "tricks" the binary classifier. And it is formally defined as false positive rate (FPR = FP/N), where FP is the count of synthetic sequences misclassified as authentic, and N is the count of all negatives, i.e., synthetic sequences.

|  | D1 | D2 | D3 | D4 | D5 | D6_1 | D6_2 |
|---|---|---|---|---|---|---|---|
| F1 | 0.94 | 0.95 | 0.93 | 0.89 | 0.98 | 0.96 | 0.98 |
| GRU | 75.6% | 82.4% | 91.2% | 63.4% | 94.8% | **91.6%** | 87.6% |
| LSTM | 78.4% | 82.4% | 89.8% | 47.4% | 91.0% | 90.4% | 86.8% |
| Trans-AR | 72.2% | 64.4% | 90.0% | 66.2% | 80.0% | 88.4% | 82.4% |
| Trans-NAR | 65.8% | 65.8% | 84.6% | 21.6% | 81.8% | 78.8% | 72.6% |
| P-GAN | 80.0% | **90.4%** | 96.2% | 71.6% | 75.8% | 84.0% | 97.6% |
| P-GAN-M | **84.4%** | 77.2% | 97.2% | 72.6% | 95.0% | 75.6% | **98.0%** |
| P-GAN-K | 70.4% | 89.4% | **98.2%** | **87.2%** | **97.2%** | 77.0% | 94.4% |

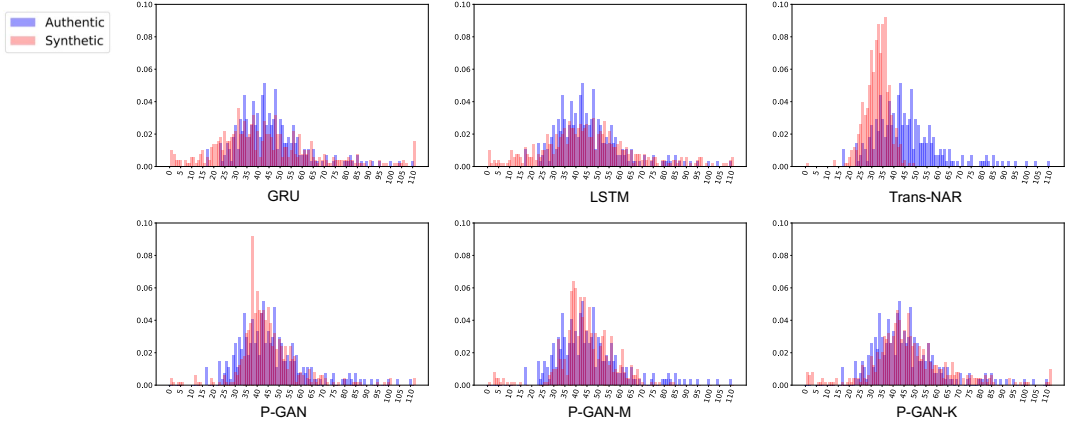

**Figure 4: The length distribution of the synthetic sequences generated by different models for dataset D1. The x-axis represents the sequence length (i.e., the number of activities), and the y-axis represents the fraction of each length in the dataset.**

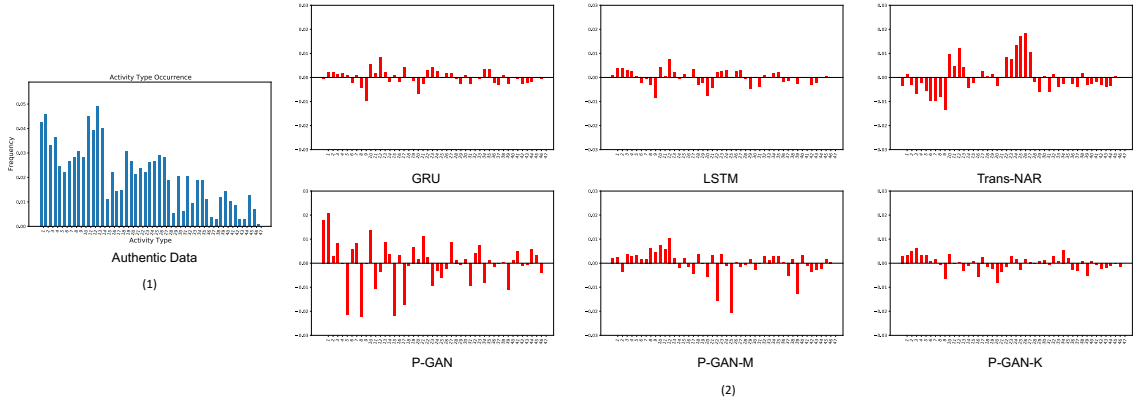

**Figure 5: (1) The activity occurrence distributions of the authentic sequences in dataset D1. (2) The difference of activity occurrence frequency between authentic and synthetic sequences generated by different models. The x-axis represents the activity types, and the y-axis represents the difference.**

From the sequence variance perspective (Table 4), we found that all models can generate variable sequences. When the data size is large (D5, D6_1), all models performed about the same. When the amount of process data was small or more complex, the GAN based models performed better in generating sequences with similar variance to the authentic data.

From the statistical results, we found that the process data complexity and the data size can impact the performance of different models. We found the GAN-based models provide more regularizations when sequences are complex and when data size is insufficient, which makes these models suitable for process data augmentation in such conditions. Autoregressive generative models are better at mimicking the observed process data.

**Supervised Learning Score.** We used the false positive rate (percentage of the synthetic sequences being classified as authentic) as the metric to measure the authenticity (Table 5). The more sequences being classified as positive, the better our process generator performs.

We found the GAN-based models have better scores than other models for most of the datasets, which means the activities in these sequences have a more reasonable order. Considering that the Transformer-based classification model can capture more intermediate dependencies between the tokens, a higher score proved that our models managed generating sequences with complex activity dependencies. We found the models with better results in activity type occurrence and sequence variance did not always get higher scores. The models capturing a good global similarity between the sequences does not mean they captured good local activity dependencies. We found that in D6_1, the GAN-based models got lower score and the baseline autoregressive models achieved higher scores, and in D6_2 when we increased the process length, the score of the baseline RNN-related models dropped and were surpassed by the GAN-based models, which means the increase of process complexity reduces the performance of RNN-related models. Our GAN-based models are more robust with different process complexity. The Transformer model also outperformed the RNN-related models because the multi-head attention mechanism achieves better performance in longer sequences.

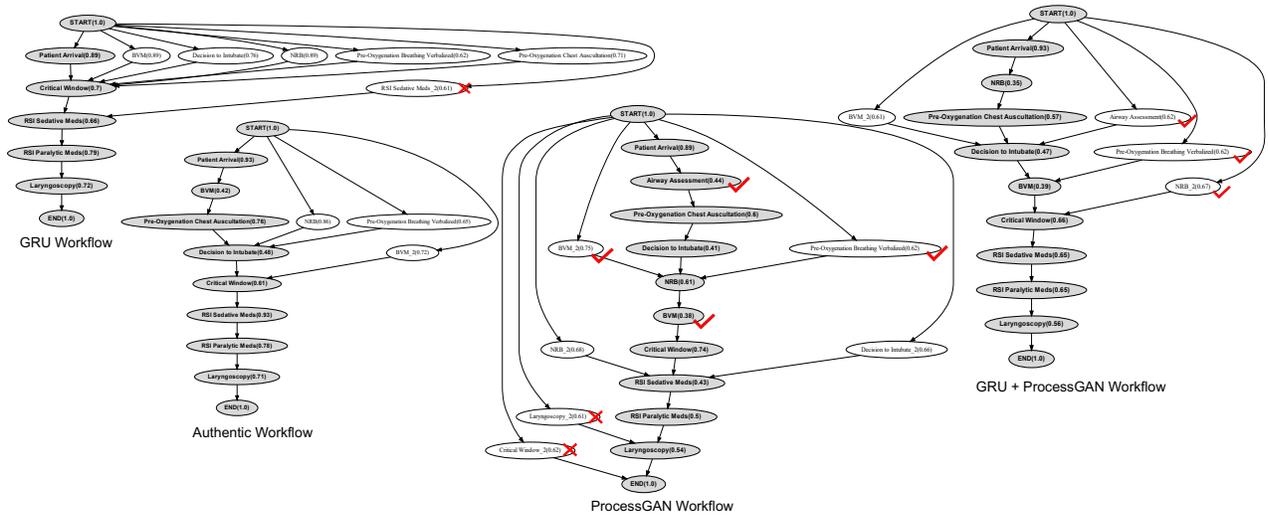

**Figure 6: The workflow diagrams of the authentic Intubation dataset (D2) and three synthetic ones discovered by the alignment method [4]. The gray ovals are the backbone activities that represent the major process, and the white ovals are the side-branch activities that represent the parallel activities. The numbers in the brackets represent the frequency of the activity. The mismatches are marked with red signs (tick: potential correct order, cross: incorrect order).**

### 4.3.2 Case Study

**Workflow Discovery.** We used the Intubation dataset (D2) as a case study to discover the workflow diagrams and further evaluate the quality of the synthetic process data. We generated workflows for the authentic traces, the GRU-generated traces and the ProcessGAN-generated traces. We also combined the synthetic traces of both models (Figure 6). In synthetic workflows, the order of "Patient Arrival", "BVM", "Decision to Intubate", "Critical Window", "RSI Sedative Meds", "RSI Paralytic Meds" and "Laryngoscopy" activities matched the authentic workflow.

We observed some discrepancies between the synthetic and authentic workflows, including different activity orders and positions. To address these, we invited medical experts to evaluate the clinical applicability of the GRU and ProcessGAN synthetic workflows:

1. The activities "NRB" and "Pre-Oxygenation Breathing Verbalized" precede "Decision to Intubate" in the authentic workflow. We observed these activities followed the "Decision to Intubate" in the ProcessGAN workflow. Our medical experts confirmed these activities could happen following the decision to intubate.
2. The "Airway Assessment" activity appears in the ProcessGAN workflow and not in the authentic workflow. According to the medical experts, airway assessment is in the correct position and routinely performed [18].
3. An additional "RSI Sedative Meds" activity spanned between the start point and the "RSI Paralytic Meds" activity in the GRU workflow. Our experts state that "RSI Sedative Meds" must immediately precede "RSI Paralytic Meds" and cannot span from the start point and "RSI Paralytic Meds".
4. Additional "Laryngoscopy" and "Critical Window" activities spanned the entire diagram in the ProcessGAN workflow. These activities cannot span the entire process based on domain knowledge.

The medical experts found that both synthetic workflows captured the major treatment steps of intubation. As RNN is not good at capturing long-range dependencies, GRU tended to generate shorter traces (Figure 4, upper left corner). GRU workflow performed well by representing the activities in correct positions of the process. From the GRU workflow (Figure 6), we observed that all the activities before "Critical Window" could be executed in parallel, and the activities after it require a logical ordering. The GRU traces fit the real clinical scenarios, although some traces have "RSI Sedative Meds" in an erroneous position.

The ProcessGAN workflow performed well in giving more detailed workflow. The medical experts liked the ProcessGAN workflow because it included longer traces and a variety of plausible pathways. Although the ProcessGAN generated more pathways that were different from the authentic traces, these pathways are valid according to expert knowledge. ProcessGAN expanded the probability of occurrence of valid pathways. It is not learning to "copy" the authentic data but learning to "imitate" the structural properties of the authentic data. It can generate more unobserved traces and augment the process datasets.

According to the ProcessGAN workflow, "Laryngoscopy" and "Critical Window" activities dispersed out of their consensus sequence position (Figure 6) each in 62% and 61% of the synthetic traces. Similarly, "RSI Sedative Meds" activity occurs in the

GRU workflow in 61% of synthetic traces dispersed out of its consensus sequence position. These occurrences represent process errors, so these traces are invalid. The invalid traces can be rejected or rectified using rules derived from domain knowledge [18]. Note that in a general case even the authentic dataset may contain process errors.

When the traces generated by both GRU and ProcessGAN models are combined, the discovered process workflow for the combined dataset was the most similar to the authentic process workflow (Figure 6). The pathways of the combined dataset workflow are all semantically valid, improving the authenticity of the synthetic dataset. Our future work will investigate the methods for training different generative models together to improve the quality of synthetic datasets.

## 5 Conclusion

Due to the privacy and confidentiality concerns, most process data cannot be shared publicly. This work tries to resolve this problem by proposing a framework for generating synthetic process data. The framework works by first learning deep representation of process data and uses the learnt model as a generative model for generating synthetic process data that can be shared publicly. We tested popular deep sequential models, i.e., GRU, LSTM, and Transformer. We also introduced ProcessGAN, a generative adversarial network of two non-autoregressive Transformers with customized loss function for process data. In addition, comprehensive evaluation methods were proposed to assess the quality of the generated process data. Our experiments show that, compared to other models, ProcessGAN can augment the existing process dataset and perform better in the cases of complex process data with small accessible data. Traditional deep sequential generative models on the other hand can generate process sequences that are more like observations. Overall, the augmented process data largely reduce the effort of data collection and can be shared publicly.

## REFERENCES

[1] Samy Bengio, Oriol Vinyals, Navdeep Jaitly and Noam Shazeer. 2015. Scheduled sampling for sequence prediction with recurrent neural networks. *arXiv preprint arXiv:1506.03099*, (2015).
[2] Zaharah A Bukhsh, Aaqib Saeed and Remco M Dijkman. 2021. ProcessTransformer: Predictive Business Process Monitoring with Transformer Network. *arXiv preprint arXiv:2104.00721*, (2021).
[3] Awatef HICHEUR Cairns, Billel Gueni, Mehdi Fhima, Andrew Cairns, Stéphane David and Nasser Khelifa. 2015. Process mining in the education domain. *International Journal on Advances in Intelligent Systems*, Vol. 8. 1 (2015), 219-232.
[4] Shuhong Chen, Sen Yang, Moliang Zhou, Randall Burd and Ivan Marsic. 2017. Process-oriented iterative multiple alignment for medical process mining. In 2017 IEEE international conference on data mining workshops (ICDMW). 438-445.
[5] Kyunghyun Cho, Bart Van Merriënboer, Dzmitry Bahdanau and Yoshua Bengio. 2014. On the properties of neural machine translation: Encoder-decoder approaches. *arXiv preprint arXiv:1409.1259*, (2014).
[6] Ian Goodfellow, Jean Pouget-Abadie, Mehdi Mirza, Bing Xu, David Warde-Farley, Sherjil Ozair, Aaron Courville and Yoshua Bengio. 2014. Generative adversarial nets. *Advances in Neural Information Processing Systems*, Vol. 27 (2014).
[7] Jiatao Gu, James Bradbury, Caiming Xiong, Victor OK Li and Richard Socher. 2017. Non-autoregressive neural machine translation. *arXiv preprint arXiv:1711.02281*, (2017).
[8] Ishaan Gulrajani, Faruk Ahmed, Martin Arjovsky, Vincent Dumoulin and Aaron Courville. 2017. Improved training of wasserstein gans. *arXiv preprint arXiv:1704.00028*, (2017).
[9] Jiaxian Guo, Sidi Lu, Han Cai, Weinan Zhang, Yong Yu and Jun Wang. 2018. Long text generation via adversarial training with leaked information. In Proceedings of the AAAI Conference on Artificial Intelligence.
[10] Sepp Hochreiter and Jürgen Schmidhuber. 1997. Long short-term memory. *Neural Computation*, Vol. 9. 8 (1997), 1735-1780.
[11] Eric Jang, Shixiang Gu and Ben Poole. 2016. Categorical reparameterization with gumbel-softmax. *arXiv preprint arXiv:1611.01144*, (2016).
[12] Tero Karras, Samuli Laine, Miika Aittala, Janne Hellsten, Jaakko Lehtinen and Timo Aila. 2020. Analyzing and improving the image quality of stylegan. In Proceedings of the IEEE/CVF Conference on Computer Vision and Pattern Recognition. 8110-8119.
[13] Jingyuan Li, Sen Yang, Shuhong Chen, Fei Tao, Ivan Marsic and Randall S Burd. 2018. Discovering interpretable medical workflow models. In 2018 IEEE International Conference on Healthcare Informatics (ICHI). 437-439.
[14] Jiwei Li, Will Monroe, Tianlin Shi, Sébastien Jean, Alan Ritter and Dan Jurafsky. 2017. Adversarial Learning for Neural Dialogue Generation. *arXiv preprint arXiv:1701.06547*, (2017).
[15] Tomas Mikolov, Ilya Sutskever, Kai Chen, Greg S Corrado and Jeff Dean. 2013. Distributed representations of words and phrases and their compositionality. In Advances in Neural Information Processing Systems. 3111-3119.
[16] Dominic A Neu, Johannes Lahann and Peter Fettke. 2021. A systematic literature review on state-of-the-art deep learning methods for process prediction. *Artificial Intelligence Review*, (2021), 1-27.
[17] Weili Nie, Nina Narodytska and Ankit Patel. 2018. Relgan: Relational generative adversarial networks for text generation. In International conference on learning representations.
[18] Karen J O'Connell, Sen Yang, Megan Cheng, Alexis B Sandler, Niall H Cochrane, JaeWon Yang, Rachel B Webman, Ivan Marsic and Randall Burd. 2019. Process conformance is associated with successful first intubation attempt and lower odds of adverse events in a paediatric emergency setting. *Emergency Medicine Journal*, Vol. 36. 9 (2019), 520-528.
[19] Marc'Aurelio Ranzato, Sumit Chopra, Michael Auli and Wojciech Zaremba. 2015. Sequence level training with recurrent neural networks. *arXiv preprint arXiv:1511.06732*, (2015).
[20] Eric Rojas, Jorge Munoz-Gama, Marcos Sepúlveda and Daniel Capurro. 2016. Process mining in healthcare: A literature review. *Journal of Biomedical Informatics*, Vol. 61 (2016), 224-236.
[21] Alex Sherstinsky. 2020. Fundamentals of recurrent neural network (RNN) and long short-term memory (LSTM) network. *Physica D: Nonlinear Phenomena*, Vol. 404 (2020), 132304.
[22] Connor Shorten and Taghi M Khoshgoftaar. 2019. A survey on image data augmentation for deep learning. *Journal of Big Data*, Vol. 6. 1 (2019), 1-48.
[23] Latanya Sweeney. 2000. Simple demographics often identify people uniquely. *Health (San Francisco)*, Vol. 671. 2000 (2000), 1-34.
[24] Farbod Taymouri, Marcello La Rosa, Sarah Erfani, Zahra Dasht Bozorgi and Ilya Verenich. 2020. Predictive business process monitoring via generative adversarial nets: The case of next event prediction. In International Conference on Business Process Management. 237-256.
[25] Wil MP Van Der Aalst, Hajo A Reijers, Anton JMM Weijters, Boudewijn F van Dongen, AK Alves De Medeiros, Minseok Song and HMW Verbeek. 2007. Business process mining: An industrial application. *Information Systems*, Vol. 32. 5 (2007), 713-732.
[26] Ashish Vaswani, Noam Shazeer, Niki Parmar, Jakob Uszkoreit, Llion Jones, Aidan N Gomez, Łukasz Kaiser and Illia Polosukhin. 2017. Attention is all you need. In Advances in Neural Information Processing Systems. 5998-6008.
[27] Zhenchen Wang, Puja Myles and Allan Tucker. 2019. Generating and evaluating synthetic UK primary care data: preserving data utility & patient privacy. In 2019 IEEE 32nd International Symposium on Computer-Based Medical Systems (CBMS). 126-131.
[28] AJMM Weijters and Wil MP van der Aalst. 2001. Process mining: discovering workflow models from event-based data. In Belgium-Netherlands Conf. on Artificial Intelligence.
[29] Sen Yang, Yichen Zhou, Yifeng Guo, Richard A Farneth, Ivan Marsic and Burd S Randall. 2017. Semi-synthetic trauma resuscitation process data generator. In 2017 IEEE International Conference on Healthcare Informatics (ICHI). 573-573.
[30] Lantao Yu, Weinan Zhang, Jun Wang and Yong Yu. 2017. Seqgan: Sequence generative adversarial nets with policy gradient. In Proceedings of the AAAI conference on artificial intelligence.
[31] Yi Yu, Abhishek Srivastava and Simon Canales. 2021. Conditional lstm-gan for melody generation from lyrics. *ACM Transactions on Multimedia Computing, Communications, and Applications (TOMM)*, Vol. 17. 1 (2021), 1-20.
[32] Yizhe Zhang, Zhe Gan and Lawrence Carin. 2016. Generating text via adversarial training. In NIPS workshop on Adversarial Training. 21-32.